# Classification of Astronomical Bodies by Efficient Layer Fine-Tuning of Deep Neural Networks

Sabeesh Ethiraj
Data Science
UpGrad Education Private Limited
Mumbai, India
sabeesh90@yahoo.co.uk

Bharath Kumar Bolla
AI & Data
Verizon, India
Bangalore, India
bolla111@gmail.com

*Abstract*— The SDSS-IV dataset contains information about various astronomical bodies such as Galaxies, Stars, and Quasars captured by observatories. Inspired by our work on deep multimodal learning, which utilized transfer learning to classify the SDSS-IV dataset, we further extended our research in the fine tuning of these architectures to study the effect in the classification scenario. Architectures such as Resnet-50, DenseNet-121 VGG-16, Xception, EfficientNetB2, MobileNetV2 and NasnetMobile have been built using layer wise fine tuning at different levels. Our findings suggest that freezing all layers with Imagenet weights and adding a final trainable layer may not be the optimal solution. Further, baseline models and models that have higher number of trainable layers performed similarly in certain architectures. Model need to be fine tuned at different levels and a specific training ratio is required for a model to be termed ideal. Different architectures had different responses to the change in the number of trainable layers w.r.t accuracies. While models such as DenseNet-121, Xception, EfficientNetB2 achieved peak accuracies that were relatively consistent with near perfect training curves, models such as Resnet-50,VGG-16, MobileNetV2 and NasnetMobile had lower, delayed peak accuracies with poorly fitting training curves. It was also found that though mobile neural networks have lesser parameters and model size, they may not always be ideal for deployment on a low computational device as they had consistently lower validation accuracies. Customized evaluation metrics such as Tuning Parameter Ratio and Tuning Layer Ratio are used for model evaluation.

I. INTRODUCTION

There have been numerous large-scale surveys that have been done to map the universe and various astronomical objects present in it. The Sloan Digital Sky Survey (SDSS) (Blanton et al., 2017) contains information captured by the observatories which include optical, spectroscopic, and photometric information, along with an array of other observations. This paper aims to evaluate transfer learning models by sequentially fine tuning the layers in a classification task. Various transfer learning architectures including mobile architectures are evaluated to identify the ideal methodology apt for these class of models. The experiments will help us gain insights on the transfer learning models and study the effect of sequential fine tuning and model performance. Models are evaluated using metrics such as validation Accuracy, Tuning Layer Ratio and Tuning Parameter Ratio. The objective of this paper is as follows:

1) To evaluate transfer learning on variety of architectures using layer wise fine tuning

2) To evaluate the models based on the training curves

II. RELATED WORK

Machine learning and Deep learning architectures are being continually designed and utilized in many large-scale astronomical surveys. Algorithms such as SKYNET that are based on Artificial neural networks have also been used in Astronomical datasets. Skynet [1] is one such neural network used in regression, classification, and clustering algorithms. A similar neural network called AstroNN [2] was designed and built specifically for astronomical surveys used to analyze spectroscopic data of the APO - Apache Point Observatory - Galactic Evolution ExperimentThe work done by [1] demonstrates the use of this network in astronomical classification.

A. *Machine Learning on SDSS Dataset*

The work done by Acharya et al. [2] in 2018 wherein the entire SDSS-3, DR-12 dataset was classified. The classification was built on the same photometric parameters you, g, i, r, and z using PySpark on Google Proc. Cloud-based computing was used in this case due to the sheer volume of the data. Models such as KNN, Support Vector Machines, and Random Forest were evaluated for their performance, with Random Forest showed the highest performance in both binary and multi-class classification. This study was later followed by a comparative evaluation study in 2020 by Petrusevich [3] on the SDSS-4 DR 14 dataset. Baseline machine learning models such as Logistic regression, Naive Bayes Classifier, Gradient Boosting, Decision Trees, and Random Forest were applied to this dataset and the enhanced version using feature engineering techniques. It was shown that these baseline machine learning models performed better than or as good as conventional deep learning models in terms of Accuracy, Precision, and Recall on both the baseline and enhanced dataset. Similar machine learning models were also used in the classification of images, as is the work done by du Buisson et al. in 2015 [6], where different images were created from the actual images of the sky at two different points in time on the transient images of the SDSS 2- survey.

B. *Deep Learning on Astronomical Classification*

Deep learning architectures have been created using photometric parameters of the SDSS dataset and passing them through a customized CNN architecture using temporal and filter convolutions [5]. Combining these models with a baseline machine learning model such as KNN and Random

Forest Classifier increased the performance of these models. Similar work was done by Khramtsov et al. in 2019 [6] on the SDSS DR 9 dataset and Galaxy Zoo2 dataset, wherein models were built using the photometric parameter and later the deep network Xception [7] was combined with a SVM to build a classification algorithm.

*C. Efficient Transfer learning*

It is a well-known fact that transfer learning models need to be fine-tuned before using them for training on any given dataset. In this regard, several innovative approaches have been proposed and experimented with state-of-the-art results. The concept of adapter modules [8] was proposed which consisted of an additional network connected to the parent architecture in series and in parallel. The architecture was based on the principle of sharing of parameters and that the additional adapter module would learn the difference in features. Adapters were used to integrate both at lower and upper layers even though lower layers were responsible only for extracting low level features but however it was shown that a combination a low level and higher level integration resulted in better model performance than just baseline fine tuning of the models or high level integration of adapters. In order to learn across multiple domains the concept of residual adapters [9]were later introduced. These adapters contain parameters that are shared across multiple domains. The learning here was done across newer domains without forgetting the existing older learning. A similar work was done in the field of NLP wherein the adapter module was incorporated in the BERT framework[10]. The literature mentioned above involved addition of extra models resulting in an increase in the number of parameters. However, the work done by [11] introduced the concept of Spot tune which involved selective fine tuning of certain layers that are specific to a particular input image or instance. The study showed that the accuracies achieved this way were higher than the conventional fine-tuning approach.

III. METHODOLOGY

The fourth phase of the Sloan Digital Sky Survey (SDSS) survey, DR 16 release consists of six types of data namely, images,optical-spectra,infrared-spectra (APOGEE/APOGEE-2), IFU spectra (MaNGA), stellar library spectra (MaStar) and catalog data (parameters such as magnitudes and redshifts obtained from spectra). The dataset used in this research has taken the image data for the evaluation of various transfer learning models such as Resnet50, VGG16, Xception, EfficientNetB2, DenseNet121 and light weight architectures such as MobileNetV2 and NasnetMobile.

*A. Description of Dataset*

The images are obtained using the tabular data which contains the Ra and Dec values which represent the location of the object in the sky. A scale of 0.1 would focus on the target at the center of the image thus eliminating any artifacts present in the dataset. The dimension of the images downloaded are 2048x2048x3. A subset of the total dataset which consists of 1000 data points are used in this study. The data is split into a ratio of 70:30 into the train and the validation set. Figure 1 shows the three different classes of images present in our dataset.

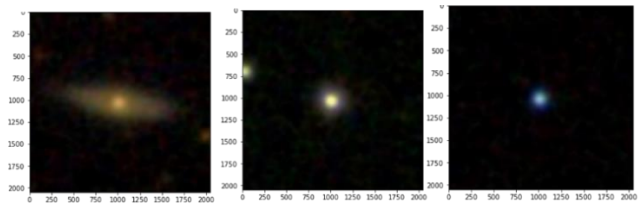

*Fig 1. Galaxies, Stars, and Quasars*

*B. Data Preprocessing*

The target variable is identified from the tabular data. A dataset is created using TensorFlow consisting of images as the training features and a categorical variable with three different categories as the target variable. The images are resized from 2048x2048x3 to 512x512x3 to reduce the number of training parameters as higher image size results in a higher dimension of the output channels.

*C. Class Imbalance*

Initial metadata analysis reveals a class imbalance in the dataset where the Quasar Class is heavily imbalanced. The class imbalance is mitigated using appropriate class imbalancing techniques such as incorporation of class weights during model training. The distribution of the target variables in the dataset is shown in Table 1

*Table 1. Class Imbalance distribution*

| Class | Total | % of Distribution |
|---|---|---|
| Galaxy | 440 | 44.0 |
| QSO | 62 | 6.2 |
| Star | 498 | 49.8 |

*D. Transfer Learning architectures*

In this study, we used Resnet50, VGG16, DenseNet121, EfficientNetB2, Xception, and mobile architectures such as MobileNetV2 and NasnetMobile for classification and evaluation of transfer learning using Imagenet weights. The original weights trained on ImageNet are used as the baseline weights before fine-tuning. The original softmax layer has been replaced by a custom softmax output with a 3 class classification (Figure 3). No augmentation techniques were used in the training algorithm. The summary of the architectures with the number of layers and the parameters is shown in Table 2. It is seen that Resnet50 has the highest number of parameters among the older architectures while mobile architectures such as MobileNetV2 and NasNet mobile have the least number of parameters.

*Table 2. Transfer learning - Parameters and Layer configuration*

| Transfer Learning Model | Layers | Parameters |
|---|---|---|
| Resnet50 | 176 | 23,593,859 |
| Xception | 133 | 20,867,627 |
| VGG16 | 20 | 14,716,227 |
| EfficientNet B2 | 340 | 7,772,796 |
| DenseNet121 | 428 | 7,040,579 |
| NasNetMobile | 770 | 4,271,830 |
| MobileNetV2 | 155 | 2,260,546 |

Table 3. Parameter-layer configuration of Various Architectures

| Base Model | Resnet50 P | Resnet50 L | VGG16 P | VGG16 L | Densenet121 P | Densenet121 L | Xception P | Xception L | EfficientNet2 P | EfficientNet2 L | MobileNetV2 P | MobileNetV2 L | NasNetMobile P | NasNetMobile L |
|---|---|---|---|---|---|---|---|---|---|---|---|---|---|---|
| BM0 | 6.1K | 0 | 1.5K | 0 | 3.07K | 0 | 6.14K | 0 | 4.22K | 0 | 3.8K | 0 | 3.1K | 0 |
| BM1 | 1.06M | 5 | 1.5K | 1 | 41.9K | 5 | 6.14K | 2 | 7.04K | 3 | 3.8K | 1 | 3.5K | 10 |
| BM2 | 3.42M | 10 | 1.5K | 2 | 171K | 10 | 3.16M | 4 | 502K | 6 | 6.4K | 3 | 137K | 20 |
| BM3 | 5.52M | 15 | 2.36M | 3 | 331K | 15 | 3.17M | 6 | 1.24M | 9 | 416K | 5 | 174K | 30 |
| BM4 | 7.88M | 20 | 4.72M | 4 | 369K | 20 | 4.75M | 8 | 1.62M | 12 | 723K | 7 | 343K | 40 |
| BM5 | 8.94M | 25 | 7.08M | 5 | 490K | 25 | 4.75M | 10 | 1.62M | 15 | 734K | 10 | 715K | 50 |
| BM6 | 14.4M | 30 | 7.08M | 6 | 642K | 30 | 5.50M | 12 | 1.64M | 18 | 1.04M | 15 | 717K | 60 |
| BM7 | 14.9M | 35 | 9.44M | 7 | 686K | 35 | 6.25M | 14 | 2.83M | 21 | 1.05M | 20 | 886K | 70 |
| BM8 | 15.8M | 40 | 11.8M | 8 | 830K | 40 | 6.79M | 16 | 2.96M | 24 | 1.36M | 25 | 1.02M | 80 |
| BM9 | 16.1M | 45 | 12.9M | 9 | 939K | 45 | 6.79M | 18 | 2.96M | 27 | 1.52M | 30 | 1.42M | 90 |
| BM10 | NA | NA | 12.9M | 10 | NA | NA | 7.33M | 20 | 2.97M | 30 | NA | NA | NA | NA |
| BM11 | NA | NA | 13.5M | 11 | NA | NA | 7.33M | 22 | 3.23M | 33 | NA | NA | NA | NA |
| BM12 | NA | NA | 14.1M | 12 | NA | NA | 7.87M | 24 | 3.49M | 36 | NA | NA | NA | NA |
| BM13 | NA | NA | 14.4M | 13 | NA | NA | 8.40M | 26 | 3.62M | 39 | NA | NA | NA | NA |

*E. Layer wise Fine-tuning*

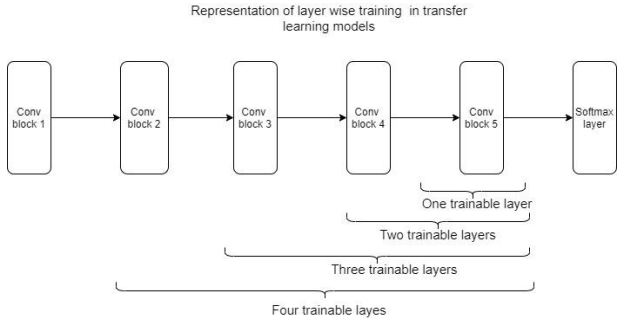

Fig 2. Tuning of Layer training - Representative diagram

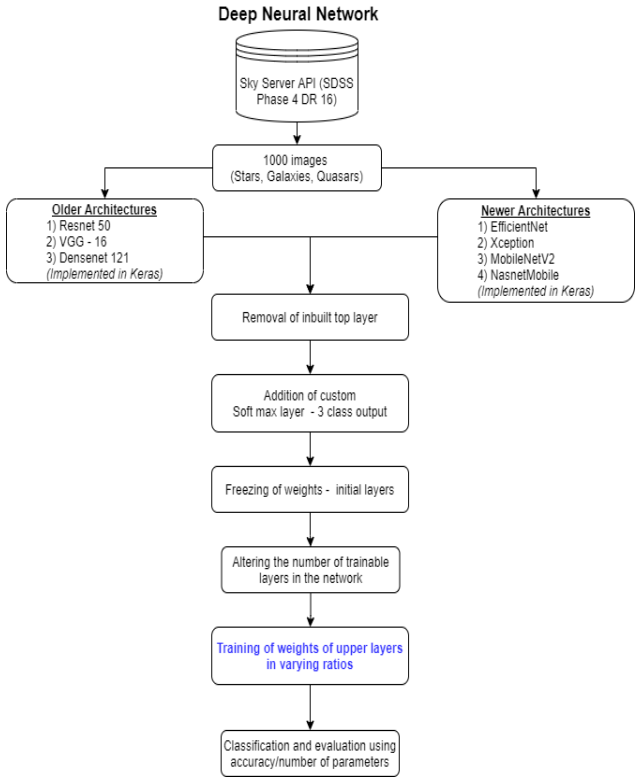

Fig 3. Transfer Learning Layer Tuning

Experiments have been done by gradually increasing the ratio of trainable layers and parameters as shown in Figure 2. The optimal number of trainable layers for a particular architecture at which the highest accuracy is obtained is considered the best performing model. The details of parameters and layers tuned are described in Table 3.

*F. Loss Functions*

The loss function used here is the Categorical Cross-Entropy loss as this is a multi-class classification problem. The following equation defines categorical cross entropy loss.

$$CE = -\sum_{i}^{C} t_i log(s_i)$$

Equation 1. Cross Entropy Loss

*G. Evaluation Metrics*

The evaluation metrics used in this study are Recall, Accuracy, Precision, and F1 score. Layer Tuning Ratio is the ratio between the number of tuned layers to the total number of layers. Parameter tuning ratio is the ratio between the number of tuned parameters to the total number of parameters.

## IV. RESULTS

*A. Transfer Learning on Image Data*

Table 4 summarizes the performance of the baseline transfer learning models without any fine tuning.

Table 4. Baseline Transfer Learning Model performance

| Models | R | V | D | X | E | M | N |
|---|---|---|---|---|---|---|---|
| Trainable Params | 6.1K | 1.5K | 3K | 6.1K | 4.2K | 3.8K | 3.1K |
| Acc | 48 | 44 | 66 | **83.5** | 80 | 78.5 | 80 |

(*R-Resnet50, V-VGG16, D-Densenet121, X-Xception, E-EfficientNetB2, M-MobileNetV2, N-NasnetMobile)

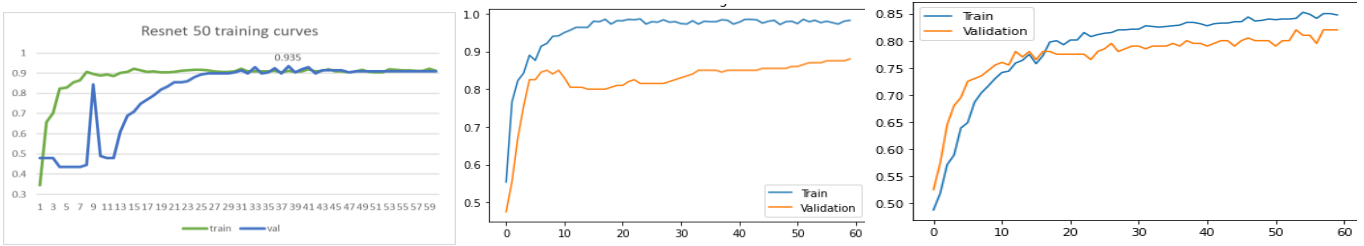

*Fig 4.Resnet, MobileNetV2 and Nasnet Training Curves - Val Accuracies*

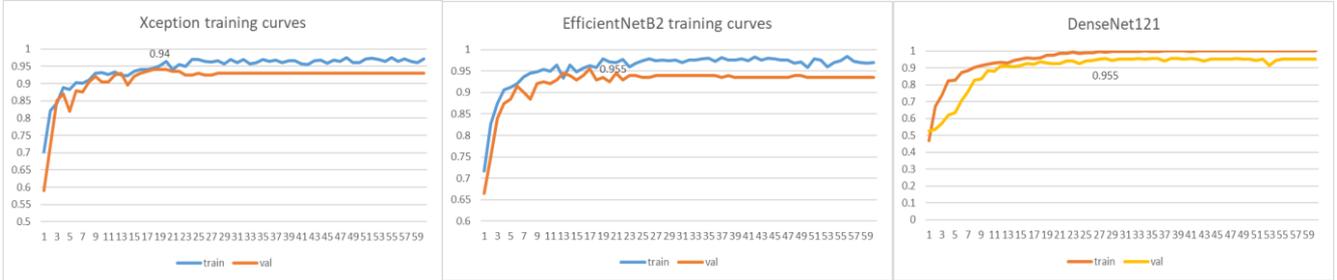

*Fig 5. Xception, EfficientNetB2, and DenseNet121 Training Curves-Val Accuracies*

### B. Baseline Models Analysis

Baseline models contain the trainable weights of only the last softmax layer. The weights of the other layers are frozen and not trained. As shown in Table 6, it is evident that baseline models without any fine-tuning perform poorly on the given dataset, with accuracy as low as 44% in the case of VGG16. It is observed that older architectures perform poorly compared to newer architectures such as Xception and EfficientNetB2, which perform significantly better as indicated by the validation accuracies. Additionally, it is observed that baseline mobile architectures outperform older architectures with accuracy comparable to that of Google's architectures. Trainable layers vs. Accuracy

Optimum model identification requires fine-tuning the number of trainable layers. As the number of trainable layers increases, the accuracy increases gradually until it either reaches a plateau with no further increase or a dip in the accuracy is observed. The models can be classified into three categories based on their number of parameters, trainable layers, and validation accuracies.

- Consistently higher accuracies of newer architectures
- Lesser Performing Older Architectures including Mobile Architectures
- Ideal Densenet121 architecture

### C. Xception and EfficientNetB2 – Consistency of Accuracies.

As illustrated in Figures 6 and 7, newer architectures such as Xception and EfficientNetB2 achieve higher accuracy even in baseline models with a single trainable layer, and the accuracy remains consistent across a range of trainable layers. Our experiments revealed that accuracy is proportional to the number of trainable layers, as EfficientNetB2 achieved 95.5 percent accuracy at 36 trainable layers and Xception achieved 95 percent accuracy at 18 trainable layers, which is slightly lower than Densenet121. Even with more trainable layers added, the accuracies remain relatively constant after attaining a peak efficiency.

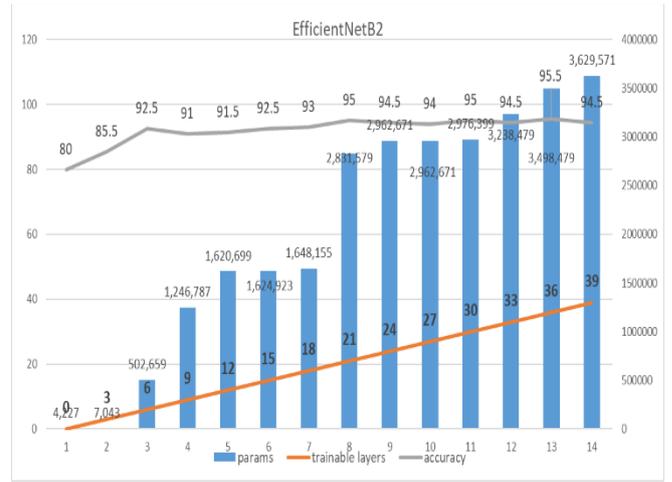

Fig 6. EfficientNetB2 Accuracy vs. Trainable Layers vs. Parameters

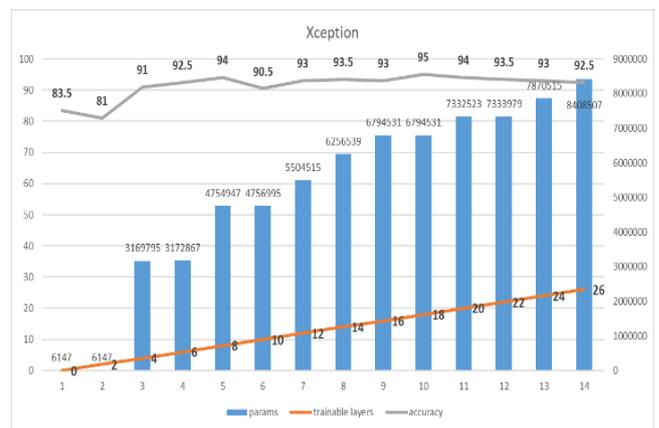

Fig 7. Xception Accuracy vs. Trainable Layers vs. Parameters

### D. Resnet50 – Dipping Accuracies

Resnet50 architecture in Figure 8 shows an increase in accuracy as the number of trainable layers increases; however, the **accuracies show a dip after 30 trainable layers**. The highest accuracy attained by this architecture is 93.5% which is slightly lower than other models.

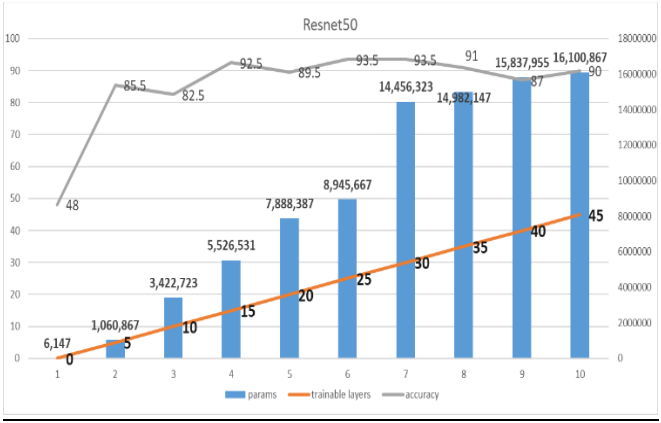

*Fig 8. Resnet50 Accuracy vs. Trainable Layers vs. Parameters*

### E. VGG – 16

After the fifth trainable layer, VGG16's validation accuracy gradually increases until it reaches a plateau of 92.5 at the tenth layer. Following that, there is no significant increase in the model's accuracy. VGG 16's accuracy graphs are shown in Figure 9.

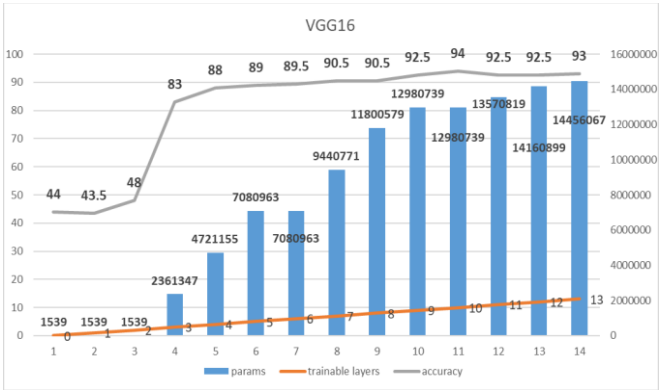

*Fig 9. VGG16 Accuracy vs Trainable Layers vs Parameters*

### F. MobileNetV2 and NasnetMobile

MobileNetV2 and NasnetMobile (Fig 10 & Fig 11) models perform similar to Resnet50 where in there is rise and dip in accuracies. The overall accuracies are slightly lower than the other architectures (90% for MobileNetV2 and 88% for NasnetMobile). However, in terms of consistency NasNetmobile performs better than MobileNetV2.

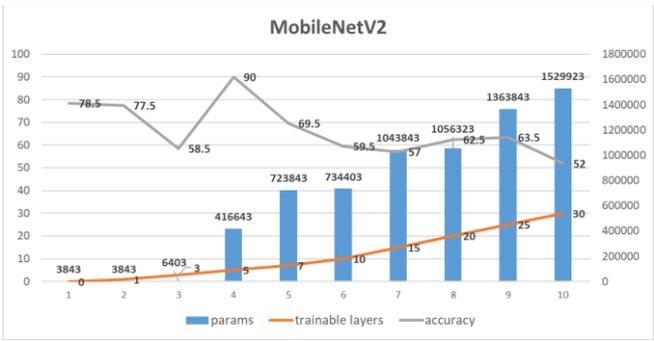

*Fig 10. MobileNetV2 Accuracy vs Trainable Layers vs Parameters*

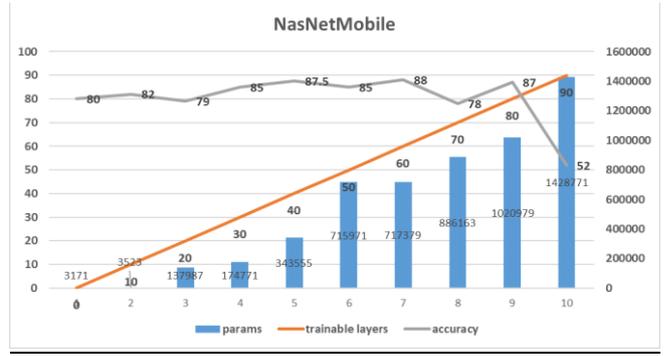

*Fig 11. NasNetMobile Accuracy vs. Trainable Layers vs. Parameters*

### G. DenseNet121 – Highest accuracy with least trainable layers

As a baseline model, DenseNet121 performs poorly in comparison to newer architectures. However, as illustrated in Figure 12, it only requires ten trainable layers to achieve the highest accuracy (95.5 percent) of all transfer learning models. The accuracy is consistently higher than the other models and remains constant (plateau effect) even after adding trainable layers.

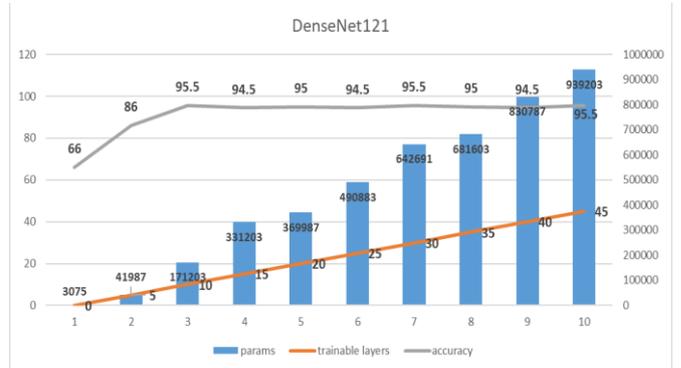

*Fig 12. DenseNet121 Accuracy vs. Trainable Layers vs. Parameters*

### H. Effect of Parameters on Accuracy

Different architectures have a different effect on the number of trainable parameters required to achieve the highest possible accuracy. DenseNet121 reached the highest accuracy while also requiring the least number of trainable parameters and layers. Even Google's Xception and EfficientNetB2 architectures are outperformed by this network. Table 7 plots the trainable parameters against the accuracy of the best-performing models for each architecture.

### I. Training curves of Transfer Learning

The training curves for the five distinct transfer learning architectures are depicted in Figure 4 and 5. Densenet121's training curves for the best trainable layers demonstrate a near-perfect fit with no overfitting/underfitting. Additionally, the curve is steeper than other architectures, with the model achieving maximum accuracy in the fewest possible epochs. Moreover, the curves of newer architectures such as Xception and EfficientNetB2 are steeper and smoother than those of Resnet50 and VGG16. Mobile architectures such as MobileNet and NasNet are similar to Resnet50 and VGG16 as they exhibit a much slower convergence rate to the optimal minima.

*Table 5. Trainable Parameters vs. Highest Accuracy*

| Model | Layers Trained | Parameters Tuned | Layer Tuning Ratio | ParameterTuning ratio | Acc |
|---|---|---|---|---|---|
| **DenseNet121** | **10** | **171,203** | **0.02** | **0.02** | **95.5%** |
| EfficientNetB2 | 36 | 3,498,479 | 0.10 | 0.45 | 95.5% |
| Xception | 18 | 6,794,531 | 0.13 | 0.32 | 95% |
| Resnet50 | 25 | 8,945,667 | 0.14 | 0.37 | 93.5% |
| VGG16 | 10 | 12,980,739 | 0.5 | 0.88 | 94% |
| MobileNetV2 | 5 | 416,643 | 0.03 | 0.18 | 90% |
| NasNetMobile | 60 | 717,379 | 0.07 | 0.16 | 88% |

*J. Layer Tuning Ratio / Parameter Tuning Ratio*

From Table 5 it is evident that DenseNet121 has the least layer tuning ratio (0.02) and the least parameter tuning ratio (0.02), indicating that the architecture achieves the peak accuracy with just minimal tuning. VGG16 has the highest ratio amongst all the architecture (0.5 & 0.88) for layer and parameter fine tuning respectively. This coupled with architectural consistency in attaining peak accuracies establishes the supremacy of the Densenet121 architecture.

## V. CONCLUSION

Straightforward transfer learning in Deep Learning should not be implemented with the transferred weights to achieve better results. Though the baseline models broadly perform well on a given dataset, fine-tuning of these models is required, specific to the dataset being used. The outcome and observations of our work can be summarized as below.

- Contemporary architectures, such as Xception and EfficinetNetB2 (Google's architectures), demonstrated consistency in terms of increased accuracy regardless of the number of layers trained in the network.
- Older architectures such as Resnet50 and VGG16 displayed an initial increase in accuracy followed by a decline as the number of trainable layers increased.
- Densenet121 architecture resulted in the highest and consistent performance among all architectures with the least layer and parameter tuning ratio.
- Mobile Architectures achieved validation accuracies comparable to Resnet50 and VGG16; however, they are lighter and have a smaller model size, and their applicability must be determined on a case-by-case basis.
- The training curves of newer architectures, including Densenet121, showed better, smoother, and faster convergence than Resnet and VGG16.

While the findings above are specific to this dataset, they may be generalizable to other large datasets, and additional research can validate the effect of layer tuning to achieve peak performance. Future work in this aspect may also be conducted by performing block wise fine-tuning instead of layer wise fine-tuning. Research into specific variations in these architectures needs to be done that can add interpretability to the findings of this research.